\documentclass[10pt, a4paper]{article}
\usepackage{lrec2022} 
\usepackage{multibib}
\newcites{languageresource}{Language Resources}
\usepackage{graphicx}
\usepackage{subfigure}
\usepackage{tabularx}
\usepackage{soul}
\usepackage{caption}
\usepackage{float}
\usepackage{titlesec}
\titleformat{\section}{\normalfont\large\bfseries\center}{\thesection.}{1em}{}
\titleformat{\subsection}{\normalfont\SmallTitleFont\bfseries\raggedright}{\thesubsection.}{1em}{}
\titleformat{\subsubsection}{\normalfont\normalsize\bfseries\raggedright}{\thesubsubsection.}{1em}{}
\renewcommand\thesection{\arabic{section}}
\renewcommand\thesubsection{\thesection.\arabic{subsection}}
\renewcommand\thesubsubsection{\thesubsection.\arabic{subsubsection}}

\usepackage{epstopdf}
\usepackage[utf8]{inputenc}

\usepackage{hyperref}
\usepackage{xstring}

\usepackage{color}

\usepackage{caption}

\title{Open Terminology Management and Sharing Toolkit for Federation of Terminology Databases}

\name{Andis Lagzdiņš$^{\dagger}$, Uldis Siliņš$^{\dagger}$, Mārcis Pinnis$^\dagger$$^{\ddagger}$, Toms Bergmanis$^\dagger$$^{\ddagger}$,\\  
{\bf \large Artūrs Vasiļevskis$^{\dagger}$ and Andrejs Vasiļjevs$^\dagger$$^{\ddagger}$}}

\address{  $^\dagger$Tilde / Vienības gatve 75A, Riga, Latvia \\
  $^\ddagger$Faculty of Computing, University of Latvia / Raiņa bulv.  19, Riga, Latvia\\ 
             \{name.surname\}@tilde.lv\\}

\abstract{
Consolidated access to current and reliable terms from different subject fields and languages is necessary for content creators and translators. Terminology is also needed in AI applications such as machine translation, speech recognition, information extraction, and other natural language processing tools.
In this work, we facilitate standards-based sharing and management of terminology resources by providing an open terminology management solution -- the EuroTermBank Toolkit. 
It allows organisations to manage and search their terms, create term collections, and share them within and outside the organisation by participating in the network of federated databases.
The data curated in the federated databases are automatically shared with EuroTermBank, the largest multilingual terminology resource in Europe, allowing translators and language service providers as well as researchers and students to access terminology resources in their most current version.
\\ \newline  }

\begin{document}

\maketitleabstract

\section{Introduction}

Language evolves: new words are coined, existing words change their meaning, and some even become unused. New concepts and terms that denote them are created every day, but many older concepts and their denotations rapidly become obsolete. Consequently, terminological data become obsolete over time if not regularly updated.
Individual term collections are usually maintained by the respective institution, such as an industrial company, an academic centre, or a public administration.
Still, many institutions lack a proper terminology management system and struggle to maintain their terms current.
This has practical and financial consequences as consolidated access to current and reliable terms from different sources is necessary not only for content creators and translators but also for artificial intelligence (AI) applications.

Terminology management is even more challenging for termbanks that provide access to term collections aggregated from different institutions.  Although terminology work benefits from a rigorous standardisation process and essential standards developed by ISO TC37 \cite{vasiljevs2006terminology}, insufficient supporting tools and infrastructure as well as different terminology management practices (including what data in what format is being stored) that are in place across Europe are factors that hinder terminology data sharing in a timely fashion \cite{gornostay2010terminology}.\\

\begin{figure}
  \includegraphics[width=\linewidth]{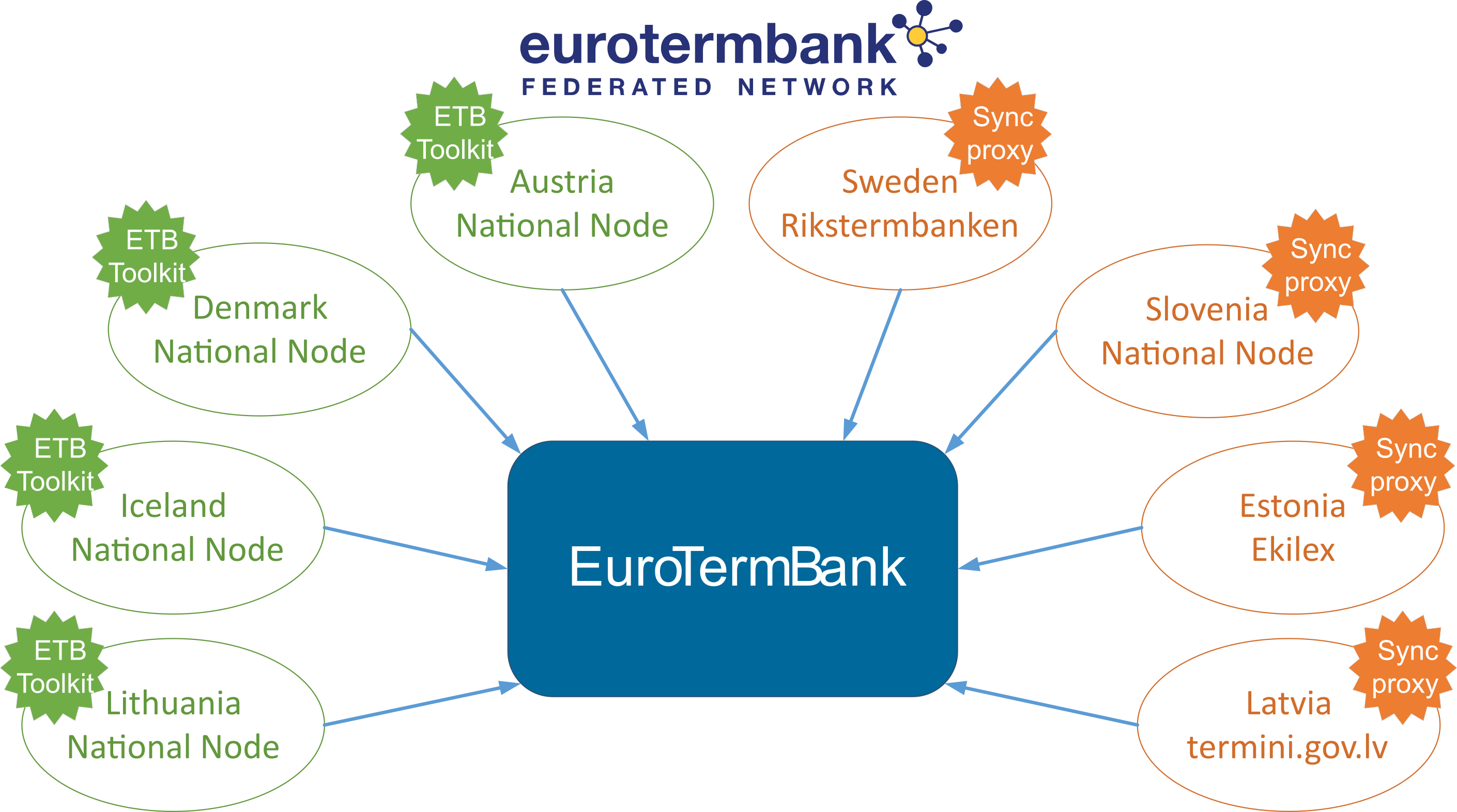}
  \caption{Federated nodes linked to the EuroTermBank Federated Network.}
  \label{fig:nodes}
\end{figure}

\begin{figure*}
\center
  \includegraphics[width=0.75\linewidth]{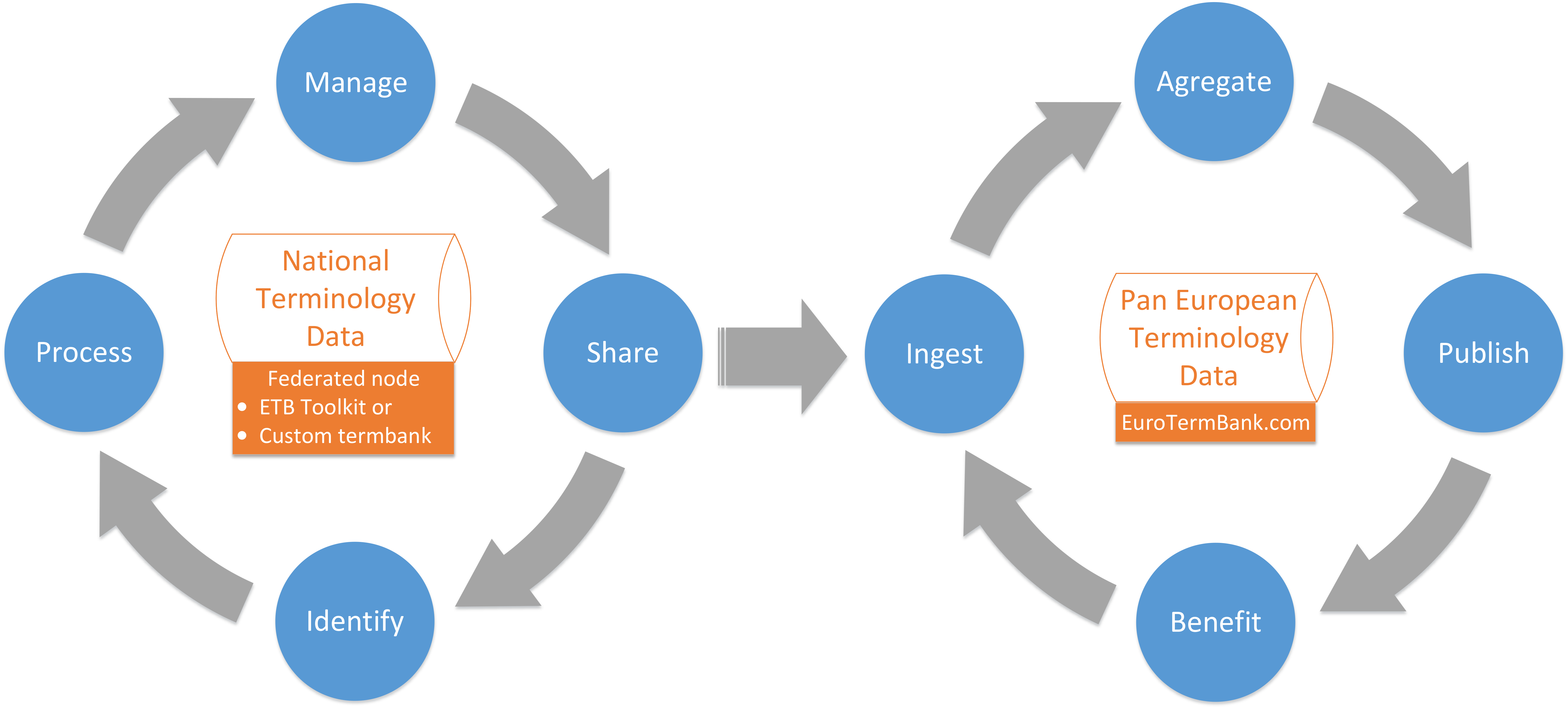}
  \caption{A conceptual depiction of EuroTermBank Federated Network.}
  \label{fig:concept}
\end{figure*}

In this paper, we describe a solution to these challenges that was created for the largest aggregation of European terminology resources, namely EuroTermBank\footnote{\url{https://www.eurotermbank.com/}} \cite{vasiljevs2008eurotermbank} and institutions participating in the EuroTermBank Federated Network. 
We present the EuroTermBank Toolkit (ETBT), an open terminology management toolkit for the EuroTermBank Federated Network that allows organisations to manage their term collections and share them within and outside the organisation.\\
The motivation of this work was to support other initiatives of natural language processing (NLP) like automated text and speech translation with reliable terminology.
In the following subsections, we briefly describe the application of terminology in these fields, then provide a short overview of EuroTermBank and the federated approach to terminology consolidation. Then, we continue with a description of the EuroTermBank Toolkit, its functionality and architecture, and the current state of the EuroTermBank Federated Network by providing statistics of terminology resources available within the network and institutions hosting federated nodes.

\subsection{Applications of Terminology in Natural Language Processing}
While terminology in NLP is sometimes considered in a monolingual setting, most of its applications are related to multilingualism and translation. Terminology data has been proven to boost the quality of machine translation in the past \cite{pinnis2015terminologijas} and has been helpful in the work of professional translators via computer-aided translation software \cite{arcan2014enhancing,arcan2017leveraging,verplaetse2019surveying}. As of relatively recently, there has been a plethora of research on terminology integration in modern machine translation systems based on artificial neural networks \cite{de2018neural,dinu2019training,DBLP:journals/corr/abs-2106-12398,bergmanis2021facilitating,exel-etal-2020-terminology,exel-etal-2020-terminology,wang2022integrating}. The sheer volume of research on terminology integration in modern machine translation systems indicates a great interest from the industry of language service providers. Similar trends can be observed with the development of machine translation of speech \cite{cross2018end,di2019adapting,vydana2021jointly} and video subtitles \cite{matusov2019customizing,siekmeier-etal-2021-tag,schioppa-etal-2021-controlling}, for which there is also a growing need for correct translation of terminology \cite{gaido2021moby}.\\
Currently, however, the use of terminology in machine translation is not hindered by the lack of technology but rather by the lack of high-quality terminology data. Unlike statistical machine translation systems, which were robust to noise that is present in training data, the modern generation machine translation systems are susceptible to poor quality training data \cite{belinkov2018synthetic,khayrallah2018impact}. The same also applies to the quality of terminology data, which is often created by humans for humans only. Data created for human consumption is often unsuitable for machines as it contains irregularities that render it machine-unreadable \cite{bergmanis2021dynamic}---findings which yet again emphasise the importance of standardised practices for the curation of machine-readable terminology data.

\subsection{Overview of EuroTermBank}
The objective of the work on EuroTermBank is to contribute to the advancement of the terminology infrastructure in all member countries of the European Union (EU)  \cite{henriksen2006eurotermbank}.
The difference between EuroTermBank and other European terminology databases, such as the Interactive Terminology for Europe\footnote{\url{https://iate.europa.eu/}}\footnote{IATE was originally named as the Inter-Agency Terminology Exchange.} (IATE) \cite{johnson2000iate}, is in their primary objectives.
Although widely used by translators across Europe, the primary goal of the IATE database, for example, is to serve agencies and institutions of the EU by creating a centralised terminology platform for their translation needs. 
Thus, while IATE consolidates term collections of EU institutions, EuroTermBank is a collection of term collections of EU and many national and other institutions. 
As a result, the terminological data assembled in EuroTermBank is not created and managed by a single community but rather in a distributed fashion, often even by geographically focused working groups.
The main stakeholders in maintaining the content of EuroTermBank  are public institutions dealing with national or international terminology work. Examples are the State Language Centre of Latvia and the Institute of the Estonian Language, which coordinate terminology work in Latvia and Estonia. Other examples include the Institute of the Lithuanian Language, the University of Copenhagen, the Culture Information Systems Centre of Latvia, the Árni Magnússon Institute for Icelandic Studies, the Jožef Stefan Institute, the International Network for Terminology -- TermNet, the Swedish Institute of Standards and the Institute for Language and Folklore.

These institutions continuously maintain their terminology resources, meaning that the terminology may be added, altered, and even discarded from the local term collections at any moment, and thus the terminology may change constantly.
This poses a challenge for EuroTermBank that aggregates the terminology resources of these institutions.
If terminology keeps changing, there needs to be an automated process that ensures the currentness of terminological data in the global terminological databases.
\begin{figure}[t]
  \includegraphics[width=\linewidth]{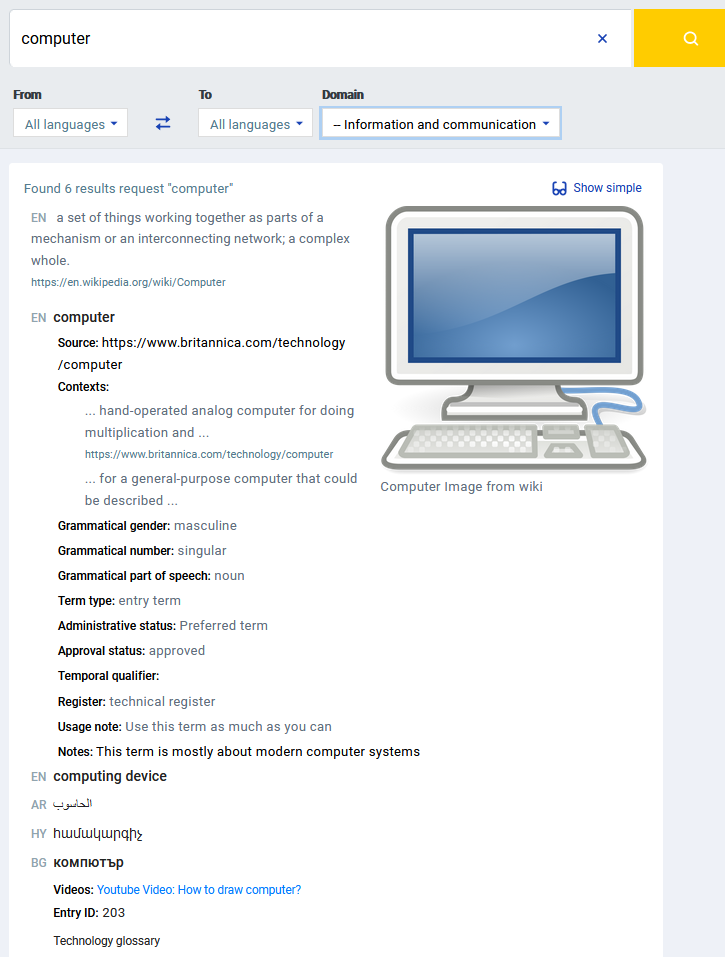}
  \caption{Term search view of ETBT.}
  \label{fig:term_search}
\end{figure}
\subsection{Federated Approach in Terminology Consolidation}
The necessity to move away from a single, isolated data bank towards a multi-bank environment was suggested by \newcite{cabre1999terminology}, who proposed simultaneously accessing several data banks that are all integrated into an overall working structure that includes not only the databases but also other computerised tools and resources. The notion of the collection of cooperating database systems that are autonomous and possibly heterogeneous has been proposed before \cite{sheth1990federated}. However, it is  \newcite{galinski2007new} who foresees the federation of term banks as a new concept in linking portals and data repositories that will go far beyond the establishment of pointers or links towards the level of exchangeability and semantic interoperability of data and data structures.  

\begin{figure}[t]
  \includegraphics[width=\linewidth]{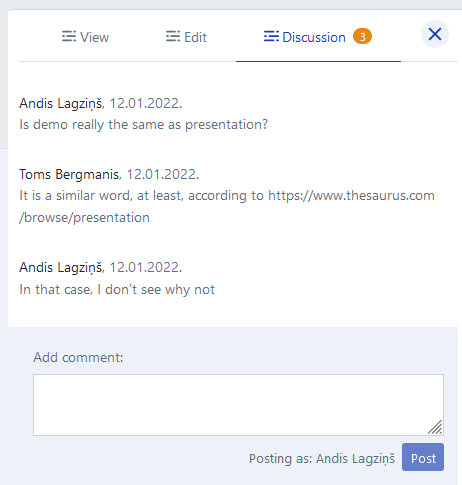}
  \caption{Term discussion view of ETBT.}
  \label{fig:discussion}
\end{figure}

A federated approach to consolidate distributed terminology resources was foreseen from the very beginning of the development of EuroTermBank \cite{vasiljevs2007consolidation}. The first implementation used distributed search queries over interlinked external termbases and aggregated returned results in a consolidated search results view. 
This implementation was eventually phased out due to serious practical drawbacks. External bases provided their results in proprietary formats that tended to change over time. Consolidation of different results into a unified structure for representation was complicated because of data format incompatibilities. There were significant delays in providing consolidated output to users due to frequent performance issues in some of the interlinked termbases.

For this reason, the federated approach presented in this paper consists of a homogeneous network of participating institutions that use unified data exchange mechanisms based on the latest versions of the TBX standard. Networked institutions either adapt the API of their existing databases to comply with the requirements of EuroTermBank Federated Network or migrate to the open EuroTermBank toolkit. Shareable data is dynamically synchronised with the central EuroTermBank database, where it is consolidated with resources coming from all participating institutions. 

\subsection{Aims of EuroTermBank Toolkit}
The ETBT aims to guarantee the \textbf{currentness of the terminological data} available at EuroTermBank by synchronising it with EuroTermBank federated nodes of organisations and institutions throughout Europe. 
The ETBT also aims to facilitate the \textbf{streamlining and standardisation of the terminology curation and sharing} practices throughout Europe, thus lowering the cost and effort required to share terminological data for both the data owners and data users. Last but not least, the open nature of the terminology management toolkit intends to \textbf{eliminate the need for non-standard processes} in terminological data sharing.
\begin{figure*}[t]
  \includegraphics[width=\linewidth]{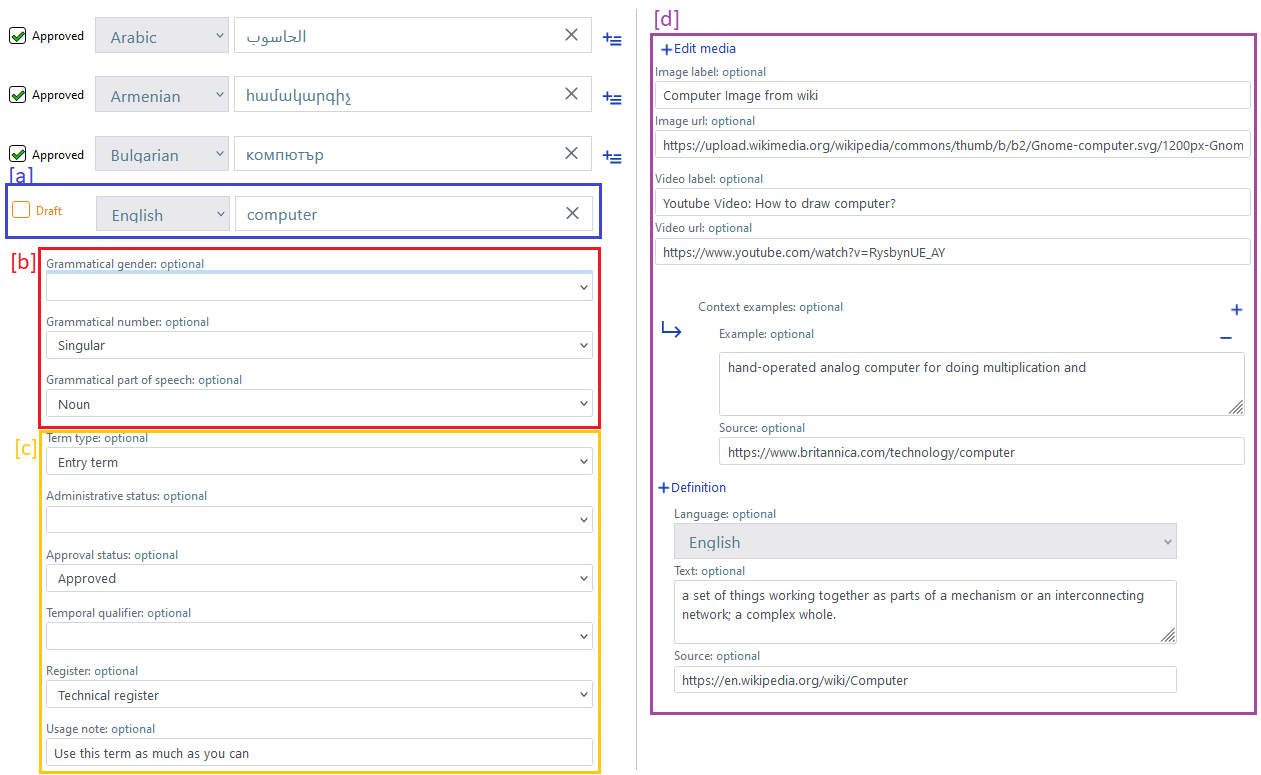}
  \caption{Term entry edit view of the ETBT. The example shows the English language side of a multilingual term.}
  \label{fig:term_edit}
\end{figure*}

\section{Concept of EuroTermBank Toolkit}
The EuroTermBank Federated Network consists of independent Federated Nodes of national, regional, or even organisational scope. 
These nodes are comprised of institutions that independently identify and coin terminology and administer it to share the resulting data with the pan-European terminology repository---EuroTermBank. EuroTermBank aggregates and publishes the terminology data to make it accessible for stakeholders in Europe and beyond. 
Figure~\ref{fig:concept} gives a conceptual view of the EuroTermBank Federated Network. 

The ETBT plays a vital role in terminology data sharing both locally and globally because most terminology work is carried out predominantly in a local setting. The ETBT facilitates standardisation and streamlining of terminology curation by offering readily available tools and infrastructure. 
For example, the ETBT is based on common standards in terminology management and sharing, such as ISO 12620 on data categories \cite{iso12620}, ISO 26162 on terminology databases \cite{iso12162}, and the TermBase eXchange (TBX) 2 standard \cite{tbx2}. 
The application of standards-based tools reduces the cost and effort of terminology curation and guarantees that the resulting terminology collections are mutually compatible, thus ensuring ease of sharing. Compatibility with the same shared standards as assured by the ETBT also enables conformity with a machine-readable data structure---an often overlooked quality for terminology, which nevertheless is paramount for terminology integration in machine translation \cite{bergmanis2021dynamic}. 

Likewise, the EuroTermBank Toolkit ensures the currentness of the terminological data available at EuroTermBank by synchronising it with EuroTermBank federated nodes of organisations and institutions throughout Europe.

\begin{figure*}[t]
  \includegraphics[width=\linewidth]{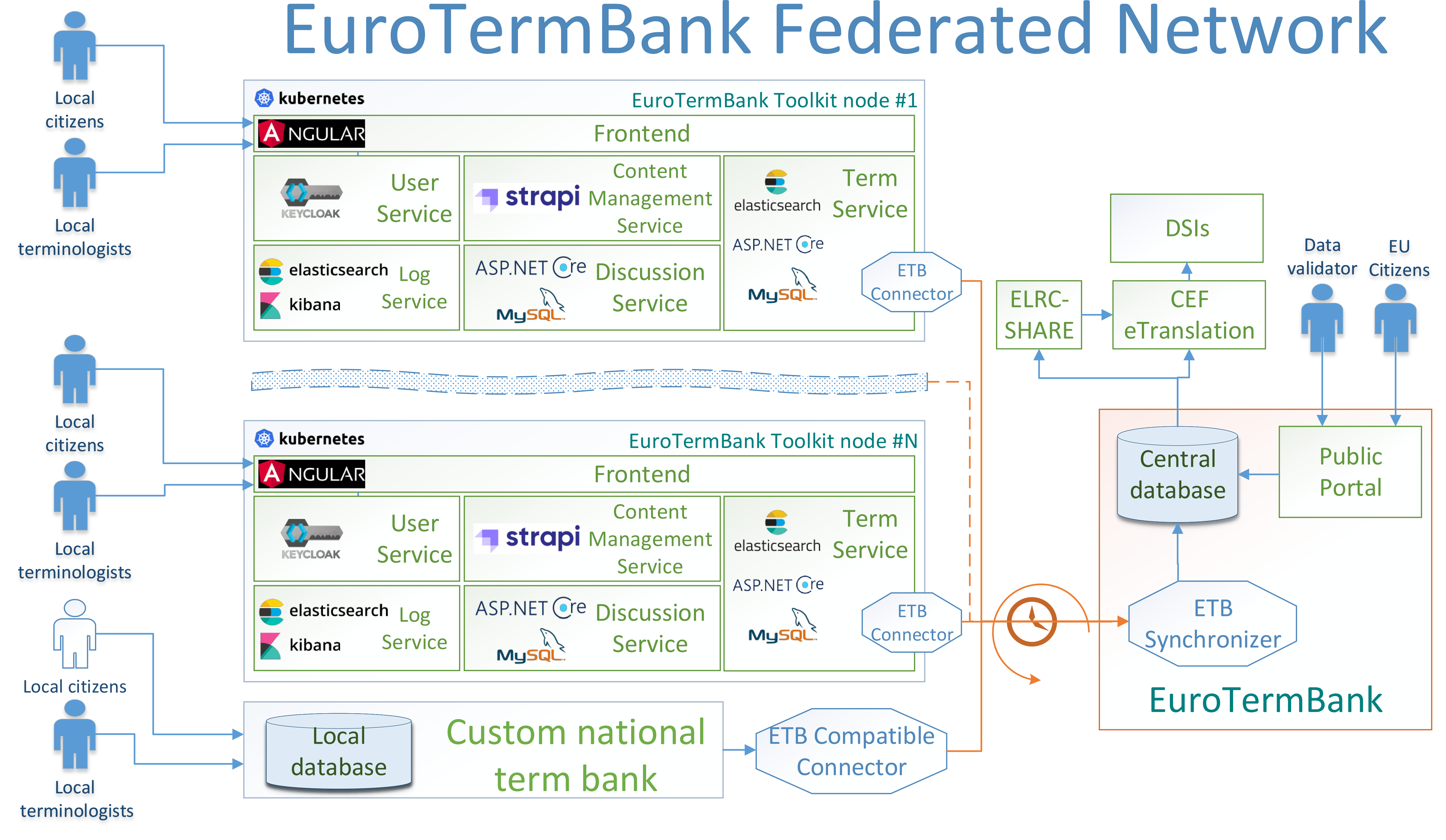}
  \caption{The architecture of the ETBT within the framework of the EuroTermBank Federated Network.}
  \label{fig:architecture}
\end{figure*}

\section{Functionality}

\paragraph{Search} Figure~\ref{fig:term_search} demonstrates the term search view of the ETBT. Terms can be searched for in the entire local database or a specific term collection or set of collections. Likewise, collections and search results can be filtered by domain and language.
\paragraph{Terminology management} Terminology data can be added in two principal ways: 1) by creating a new term entry (as well as editing an existing one) and 2) by importing an existing collection. 
Terminological information for new term entries is added by following the TBX~2 format. Besides basic data categories, such as subject field, term equivalents in different languages, definitions, and examples of how the term is used in context, information about the term's morphological properties (e.g., grammatical part of speech, number, and gender) (Figure~\ref{fig:term_edit}~b), various administrative information and usage metadata (e.g., register, type, currentness) (Figure~\ref{fig:term_edit}~c), media -- images and videos (Figure~\ref{fig:term_edit}~d) -- and other categories can be added to provide extensive information about the term. Likewise, the same information can be added for the corresponding terms in other languages, thus making the terminology collection multilingual.

Unless \textit{approved}, the term is saved as a \textit{draft} (Figure~\ref{fig:term_edit}~a), in which case it is visible only to the members of the current group and is not published. 
The import functionality supports CSV, TBX, and Excel file formats allowing to reuse already pre-existing terminology data. 

\paragraph{Terminology sharing} Term collections can be shared within a user group by adding new collaborators, or they can be exported to CSV, TBX and Excel file formats. If a term collection is made public, it is made accessible to the members of the general public through EuroTermBank.

\paragraph{Collaboration} Users can share the term candidates with collaborators, participate in discussions about the concepts and term candidates (see Figure~\ref{fig:discussion}), and approve term candidates and new entries.

\section{Architecture}
The ETBT is designed using the microservices architecture where each service can be deployed as a container using, e.g., the Kubernetes\footnote{\url{https://kubernetes.io}} container orchestration system. The architecture of the ETBT within the framework of the EuroTermBank Federated Network is depicted in Figure~\ref{fig:architecture}. The ETBT consists of six components:
\begin{itemize}
\item A \textbf{frontend application}, which provides a graphical user interface for end-users and is developed as a single page application using Angular\footnote{\url{https://angular.io}}.
\item A headless (i.e., without a graphical user interface, but with an application programming interface (API)) \textbf{content management system} (CMS) that stores static content for the frontend application. For the CMS, we use the Strapi\footnote{\url{https://strapi.io}} headless CMS.
\item A \textbf{user service}, which handles user management, authentication and authorisation. For the user service, we use the Keycloak\footnote{\url{https://www.keycloak.org}} identity and access management solution.
\item A \textbf{discussion service}, which provides functionality for terminologists to discuss individual term entries and to enable involvement in terminology work. The discussion service is built as an ASP.NET Core\footnote{\url{https://docs.microsoft.com/en-us/aspnet/core/?view=aspnetcore-6.0}} web service with an underlying MySQL\footnote{\url{https://www.mysql.com}} database.
\item A \textbf{log service} that allows to store and visualise log data. The log service utilises the Elastic Search\footnote{\url{https://www.elastic.co}} engine for data storage and retrieval and Kibana\footnote{\url{https://www.elastic.co/kibana}} for visualisation of data that is stored by Elastic Search.
\item A \textbf{term service} that provides all functionality necessary for terminology management (i.e., creation, editing, import, export, etc.), retrieval, and sharing. The term service is built as an ASP.NET Core web application with Elastic Search and an underlying MySQL database.
\end{itemize}

All terminology specified as public and thus sharable is automatically synchronised with EuroTermBank. 
The synchronisation is performed by each federated node individually. Federated nodes push changes in public term collections to EuroTermBank's Central synchronisation API. All terminological data exchange is performed using the TBX 2 data format.    

\section{Current State of the EuroTermBank Federated Network}
EuroTermBank is currently the largest centralised online terminology bank in Europe, providing access to more than 14.5 million terms from 463 collections. The EuroTermBank Federated network consortium currently consists of eight members -- four of which use a customised EuroTermBank Toolkit solution, while the other four have established a synchronisation proxy with the EuroTermBank database exchanging information with the network. These eight members represent a total of eight countries (Austria, Denmark, Estonia, Iceland, Latvia, Lithuania, Slovenia, and Sweden), which comprise academia and industry leaders in terminology and language technologies. The network's future goals are to have at least one network member in each member state of the European Union. 

\section{Conclusion}
We presented the EuroTermBank Toolkit, an open terminology management toolkit for the EuroTermBank Federated Network. The toolkit addresses the problem of outdated terminology data in shared terminology repositories by providing a standards-based infrastructure for terminology management and sharing for organisations across Europe and beyond. 
The ETBT facilitates standardisation and streamlining of terminology curation by offering readily available tools and infrastructure for collaboration and data sharing.
The common approach enabled by ETBT provides an easy to implement solution for any institution needing a standards-based tool for terminology management and data sharing. It also enables management of machine-readable data for machine translation systems and other NLP tools and facilitates data synchronisation with EuroTermBank -- the largest multilingual terminology resource in Europe. 
The instructions for the deployment of the ETBT are publicly available at: \url{https://github.com/Eurotermbank/Federated-Network-Toolkit-deployment}.

\section*{Acknowledgements}
The research leading to these results has received funding from the research project "Competence Centre of Information and Communication Technologies" of EU Structural funds, contract No. 1.2.1.1/18/A/003 signed between IT Competence Centre and Central Finance and Contracting Agency, Research No. 2.9. “Automated multilingual subtitling”.

This work was partly done within the scope of eTranslation TermBank Project (Action: 2019-EU-IA-0049) which is co-financed by the European Union's Connecting Europe Facility.

\section{Bibliographical References}\label{reference}

\bibliographystyle{lrec2022-bib}
\bibliography{lrec2022-example}



\end{document}